\documentclass[pdflatex,sn-nature]{sn-jnl}

\usepackage{graphicx}%
\usepackage{multirow}%
\usepackage{amsmath,amssymb,amsfonts}%
\usepackage{amsthm}%
\usepackage{mathrsfs}%
\usepackage[title]{appendix}%
\usepackage{xcolor}%
\usepackage{textcomp}%
\usepackage{manyfoot}%
\usepackage{booktabs}%
\usepackage{algorithm}%
\usepackage{algorithmicx}%
\usepackage{algpseudocode}%
\usepackage{listings}%
\usepackage{lmodern}%
\usepackage{subcaption}%
\usepackage{siunitx}%
\usepackage{textgreek}

\begin{document}

\title[SpecTf: Transformers Enable Data-Driven Imaging Spectroscopy Cloud Detection]{SpecTf: Transformers Enable Data-Driven Imaging Spectroscopy Cloud Detection}



\author*[1]{\fnm{Jake}~\spfx{H.}~\sur{Lee}}\email{jake.h.lee@jpl.nasa.gov}

\author[1]{\fnm{Michael}~\sur{Kiper}}\email{michael.kiper@jpl.nasa.gov}

\author[1]{\fnm{David}~\spfx{R.}~\sur{Thompson}}\email{david.r.thompson@jpl.nasa.gov}

\author[1]{\fnm{Philip}~\spfx{G.}~\sur{Brodrick}}\email{philip.brodrick@jpl.nasa.gov}

\affil[1]{\orgdiv{Jet Propulsion Laboratory}, \orgname{California Institute of Technology}, \orgaddress{\street{4800 Oak Grove Dr}, \city{Pasadena}, \postcode{91011}, \state{CA}, \country{USA}}}


\abstract{
Current and upcoming generations of visible-shortwave infrared (VSWIR) imaging spectrometers promise unprecedented capacity to quantify Earth System processes across the globe. However, reliable cloud screening remains a fundamental challenge for these instruments, where traditional spatial and temporal approaches are limited by cloud variability and limited temporal coverage. The Spectroscopic Transformer (SpecTf) addresses these challenges with a \textit{spectroscopy-specific} deep learning architecture that performs cloud detection using only spectral information (no spatial or temporal data are required). By treating spectral measurements as sequences rather than image channels, SpecTf learns fundamental physical relationships without relying on spatial context. Our experiments demonstrate that SpecTf significantly outperforms the current baseline approach implemented for the EMIT instrument, and performs comparably with other machine learning methods with orders of magnitude fewer learned parameters. Critically, we demonstrate SpecTf's inherent interpretability through its attention mechanism, revealing physically meaningful spectral features the model has learned. Finally, we present SpecTf's potential for cross-instrument generalization by applying it to a different instrument on a different platform without modifications, opening the door to instrument agnostic data driven algorithms for future imaging spectroscopy tasks.
} 




\maketitle

\newpage

\section{Introduction}


Understanding Earth's complex systems is critical for addressing global challenges like food security, biodiversity, resource management, and climate change.  Such efforts require new and refined Earth system models, an effort increasingly supported by the emerging tool of remote spectroscopy.  Remote visible-shortwave infrared (VSWIR) imaging spectrometers enable quantitative measurement of surface composition, identifying materials and their abundances using their unique signatures in the observed spectrum of reflected light.  These instruments have existed for decades in a research environment, but can now make reliable Earth-system-scale measurements thanks to the confluence of improved hardware, calibration and algorithms.  At the time of this writing, the first generation of orbital VSWIR imaging spectrometers is being deployed. Current missions such as Earth surface Mineral dust source InvesTigation (EMIT \cite{green2020emit}) and Carbon Mapper (via Tanager), alongside planned missions such as Surface Biology and Geology (SBG \cite{cawsenicholson2021nasa}) and Copernicus Hyperspectral Imaging Mission for the Environment (CHIME \cite{nieke2023chime}), will collect massive volumes of hyperspectral data. These rich datasets enable a new generation of quantitative Earth Science at scale, driven by robust retrieval algorithms of surface and atmospheric phenomena.

Clouds, though a critical element of the complete Earth System, are detrimental to optical remote sensing of Earth's surface and lower atmosphere.  VSWIR wavelengths do not penetrate optically thick clouds, and even thin clouds can impact sensitive spectroscopic measurements.  Commensurate to the challenge, a long history of algorithmic approaches exists to identify and screen out clouds, a process that is typically plagued by partially transparent clouds that distort the surface but still allow some features to come through. Typical approaches to cloud screening that are deployed in the multispectral Earth Observation world rely either on spatial \cite{zhu2012object, qiu2019fmask4, yang2019feature} or temporal \cite{hagolle2010multi, zhu2014automated, zhu2018automatic} context. Spatially dependent algorithms, however, assume a predefined structure that clouds sometimes lack, and temporal algorithms require a consistent time sequence that may not always be available.  

Imaging spectrometer data offer the potential to identify cloud properties at the pixel level with high accuracy. Despite this promise, spectroscopic methods have not yet been adopted operationally. Methods date back to the first airborne operations \cite{green1998imaging}, where the absorption depth of shallow water features discriminated even thin clouds \cite{gao1991cloud}, and continue to machine learning methods \cite{sun2020satellite, giuffrida2020cloudscout}.  Commonly, however, large-scale applications fall back to simple channel ratios or thresholds \cite{zhai2018cloud, thompson2014rapid, sandford2020global}, which have relatively low accuracies.  Why is this?  Leveraging subtle spectral features requires understanding the full spectral context, as individual features vary in magnitude and shape based on the surface and atmosphere of the observation. Global variation is enormous across locations, atmospheres, and seasons, making it exceedingly hard to develop a representative spectral training set.  Data driven methods can also be sensitive to minor systematic changes such as those that commonly occur with a periodic recalibration or a change in upstream processing approach.   Finally, different instruments sample the spectrum differently, with unique channel spacings and ranges, so models trained on raw spectroscopic data do not generally transfer to other instruments.  These factors make it difficult to guarantee model generality and have prevented widespread adoption of data-driven spectroscopic cloud screening.  Ideally, we desire an \textit{instrument agnostic} architecture that learns fundamental physical relationships rather than patterns specific to one sensor.

We address these challenges by proposing the Spectroscopic Transformer (SpecTf), a novel architecture specifically designed for imaging spectroscopy. Our main contribution shows that treating spectra as sequences rather than image channels enables efficient learning of spectral features without presuming a specific wavelength grid and without relying on spatial context. We demonstrate SpecTf's capabilities through a significantly improved cloud screening method for EMIT data, showing that spectral information alone can achieve performance only previously achieved by spatially-dependent approaches. The method's attention mechanism provides natural interpretability, revealing physically meaningful patterns learned during the cloud detection task - a crucial feature for building trust in scientific applications. Furthermore, we demonstrate SpecTf's potential for cross-instrument generalization by successfully applying trained models to different instruments, leveraging its treatment of input spectra as discrete samples of continuous spectral functions. We conclude by analyzing the method's strengths and limitations, setting the stage for future applications across imaging spectroscopy tasks.

\section{Results}

\subsection{Spectroscopic transformers treat spectra as sequences}

\begin{figure}[t]
    \centering
    \includegraphics[width=0.7\linewidth]{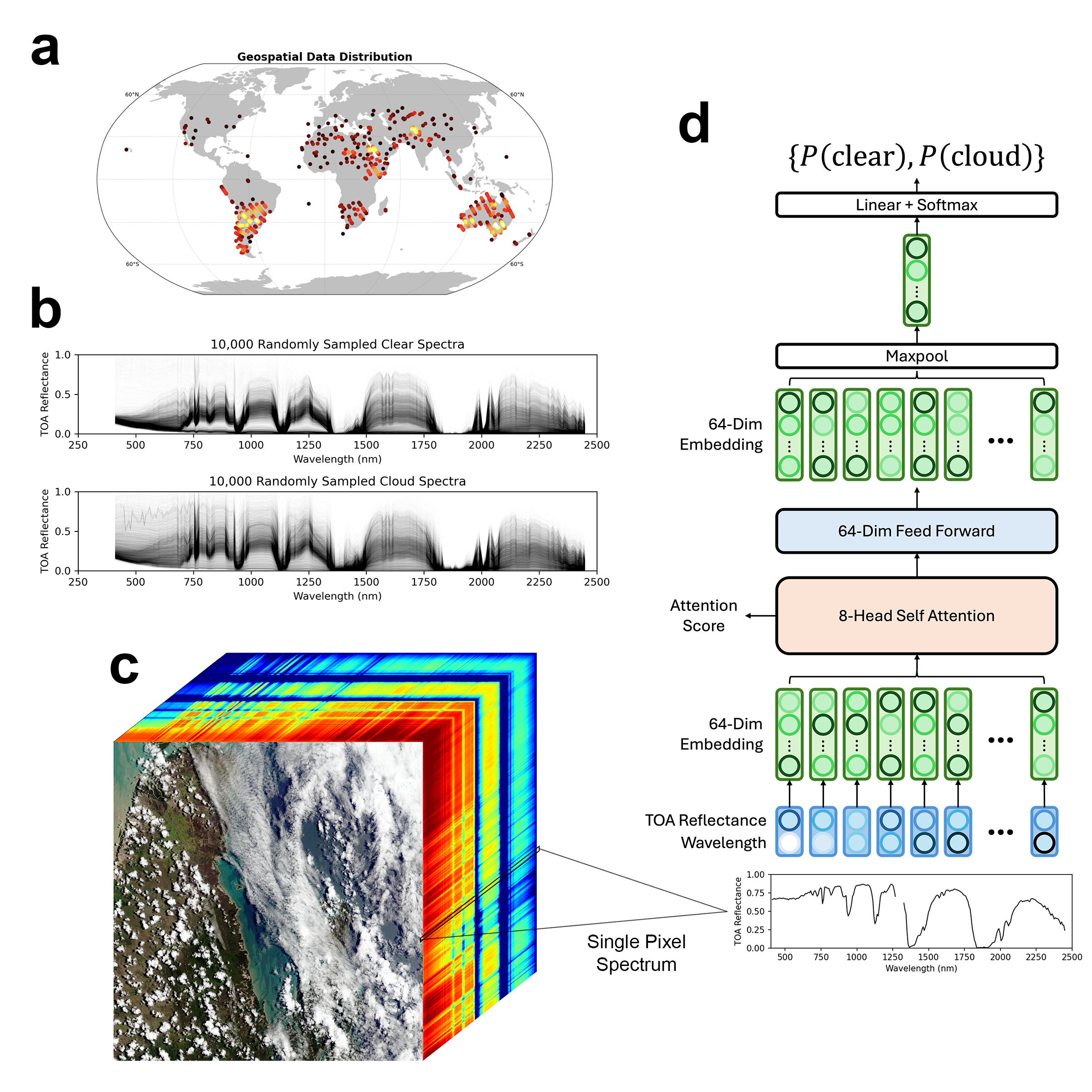}
    \caption{Overview of the SpecTf model used for cloud detection. \textbf{a}, A map of the EMIT scenes labeled for model training and validation. Brighter colors indicate density in regions where markers overlap. \textbf{b}, Visualization of a distribution of 10,000 randomly selected spectra labeled as ``clear'' and ``cloud''. \textbf{c}, a visualization of the radiance spectral cube of the EMIT scene \texttt{emit20240302t005829} and its 285 wavelengths. From each scene, individual spectra are passed to the SpecTf model, shown in \textbf{d}. SpecTf first embeds each band's wavelength and TOA reflectance into a higher-dimensional embedding. The self attention mechanism prioritizes important wavelengths relevant for cloud detection, and the feed forward, maxpool, and linear layers predict probabilities for the ``clear'' and ``cloud'' classes. This probability is thresholded to produce a binary cloud mask for downstream products.}
    \label{fig:arch}
\end{figure}

The Spectroscopic Transformer (SpecTf) is an efficient and interpretable deep-learning sequence model for cloud detection in imaging spectroscopy that shows significant generalization potential. As a pixelwise model, SpecTf gleans all of the information needed for cloud detection from each observed spectrum; by faithfully treating the consecutive channels as a discrete sequence sampled from a continuous spectrum, SpecTf is able to learn a small, well-performing model that is physically interpretable and flexible to spectra observed by different instruments without needing spatial features. 

Training the model to this ideal, as with most data-driven remote sensing models, requires a dataset with a global sampling of observations (Fig.~\ref{fig:arch}a) and accurate labels of clear and cloud-obscured pixels (Fig.~\ref{fig:arch}b). The dimensions of each standard EMIT Level 1B (L1B) radiance cube are 1242 samples (columns), 1280 lines (rows), and 285 bands (channels), as visualized in Figure~\ref{fig:arch}c. The top-of-atmosphere (TOA) reflectance is calculated from the radiance and observation geometry to normalize the spectra by the amount of incident irradiation, and the 285-dimension vector from each pixel is provided as independent inputs to the SpecTf model as a sequence.

The SpecTf architecture (Fig.~\ref{fig:arch}d) first pairs the wavelength and TOA reflectance values, encoding the position of each item in the sequence. While transformer architectures commonly utilize a relative or absolute additive cyclic positional encoding, we determined through an architecture search that concatenating the absolute wavelength best encoded the positional information for the spectral sequence. The absolute encoding is physically appropriate because each wavelength is associated with a specific photon energy; the specific interaction of matter with light at that wavelength is unique across the electromagnetic spectrum.  A fully connected layer then embeds each reflectance-wavelength pair into a higher dimensional latent space to allow for a more expressive featurization of the input. Next, the self-attention layer learns to prioritize items in the sequence that are more relevant for the cloud detection task by applying attention scores that affect the contribution of certain bands to the output. During this process, it compares each item to every other item in the sequence, identifying useful relationships for the learned task. This combined mechanism is an intuitive one for spectroscopy, where absorptions at certain wavelengths and the overall shape of the spectra indicate physical phenomena. Then, a feed forward module of dense layers learns to interpret the results for signals of cloud presence, and a maxpool layer combines signals across the sequence. Finally, a linear dense layer and a softmax layer form a classification head to produce a probability for each of the ``clear'' and ``cloud'' classes. Applying the model across every pixel in the scene produces a cloud probability map, which is thresholded to produce the final binary cloud mask.

\subsection{Evaluating model performance on EMIT scenes}

We evaluate the performance of SpecTf on a dataset of labeled scenes held-out during model training. For context, we also provide results from several reference Machine Learning (ML) models along with the current EMIT approach of using a series of specific wavelength thresholds (\cite{thompson2024atbd};  hereafter referred to as the \textit{baseline}). We also interpret the model's attention scores, and present out-of-distribution model generalization by applying the model to data from a different instrument and platform.

\begin{table}[t]
    \centering
    \begin{tabular}{l|rrrr}
         Metric & Baseline & GBT & ANN & SpecTf \\
         \hline \hline
         Learned Params. & N/A & \num{5e4} & \num{2e6} & \textbf{\num{2e4}} \\
         Binary Thresh. & N/A & $\geq 0.96$ & $\geq 0.98$ & $\geq 0.52$ \\
         \hline
         TPR        & 0.224 & 0.934 & 0.943 & \textbf{0.944} \\
         FPR        & 0.012 & 0.038 & \textbf{0.029} & 0.039 \\
         ROC AUC    & 0.606 & \textbf{0.987} & 0.957 & 0.982 \\
         \hline 
         $F_{1.0}$  & 0.363 & 0.947 & \textbf{0.956} & 0.952 \\
         $F_{0.5}$  & 0.576 & 0.955 & \textbf{0.964} & 0.957 \\
         $F_{0.25}$ & 0.796 & 0.959 & \textbf{0.968} & 0.959 \\
         $F_{0.1}$  & 0.918 & 0.960 & \textbf{0.969} & 0.960
    \end{tabular}
    \caption{Quantitative cloud detection performance of the baseline EMIT L2A cloud mask, the Gradient Boosted Tree (GBT) model, the Artificial Neural Network (ANN) model, and the Spectroscopic Transformer (SpecTf) model on the test dataset of scenes held-out during model training. For GBT, ANN, and SpecTf, models that predict probabilities, the thresholds for the best $F_1$ scores were used to calculate all binary metrics.}
    \label{tab:metrics}
\end{table}

Quantitative evaluation of a held-out dataset shows that data-driven machine learning models significantly outperform the baseline cloud mask product. The baseline (the cloud mask delivered as part of the current EMIT L2A data product) is produced by a trivariate band threshold for opaque clouds and a single band threshold for cirrus clouds \citep{thompson2024atbd, gao2002algorithm, thompson2014rapid}. The reference ML models include a gradient boosted tree (GBT) and an artificial neural network (ANN) trained on the same dataset as SpecTf.  Table~\ref{tab:metrics} shows that the GBT, ANN, and SpecTf models have a significantly higher True Positive Rate (TPR) than the baseline, at a minimal increase in the False Positive Rate (FPR). The overall improvement in trade-off, independent of the chosen binary threshold, is also reflected by the higher Receiver Operating Characteristic Area Under the Curve (ROC AUC). ML models also outperform the baseline in $F_\beta$-scores, which are harmonic means of detection precision and recall. Even when weighing precision 10 times higher than recall ($F_{0.1}$-score), ML models are better than the baseline.

\begin{figure}[t]
    \centering
    \includegraphics[width=0.7\linewidth]{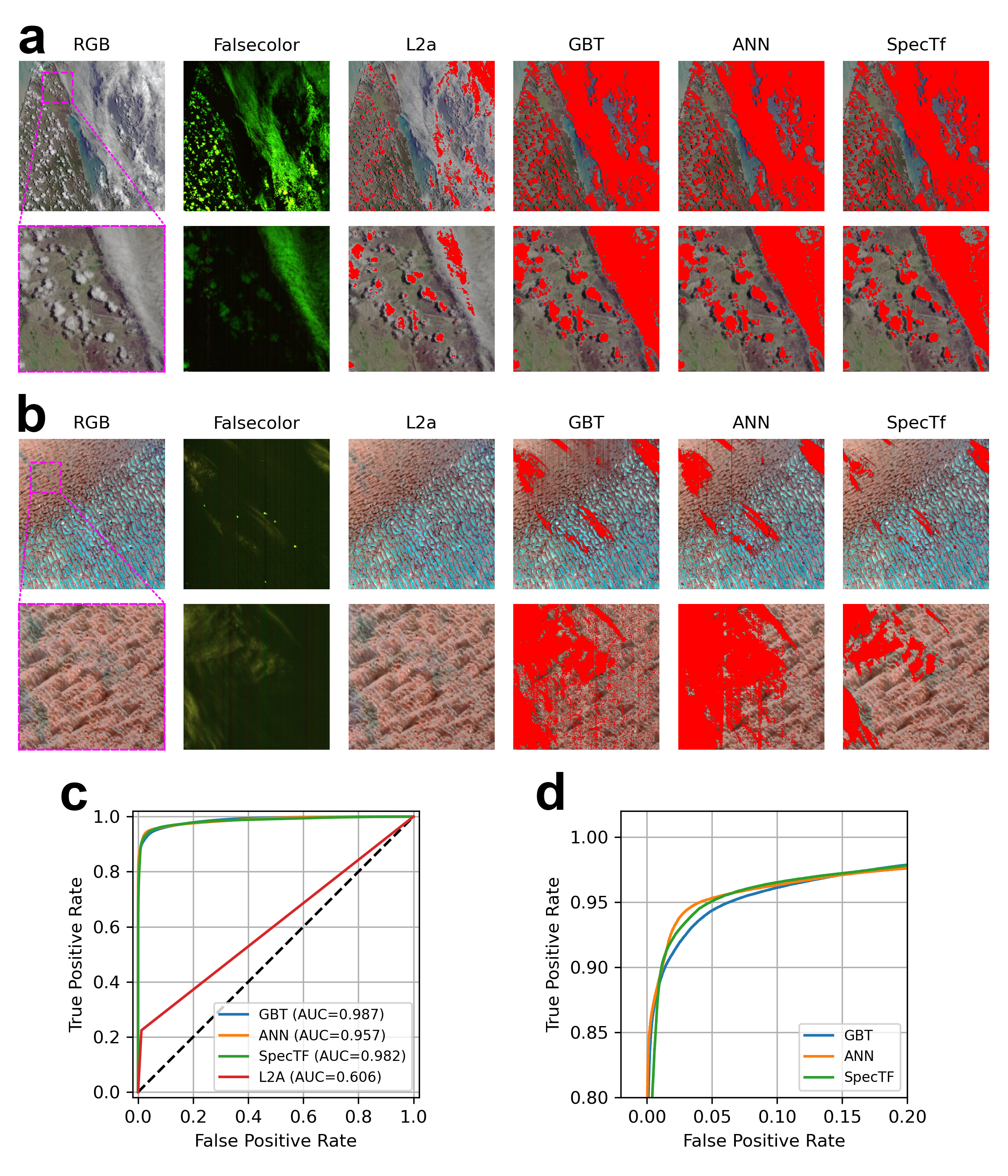}
    \caption{\textbf{a}, Examples of cloud masks (overlaid in red) produced by data-driven models on the held-out EMIT scene \texttt{emit20240302t005829}. ``RGB'' is the EMIT L1B RGB quicklook product generated from visible wavelengths, and the region outlined in magenta is magnified below. ``False color'' visualizes the 1380, 1420, and 1890 nm wavelengths for easier cloud identification. ``L2A'' is the baseline EMIT L2A cloud and cirrus mask products combined. ``GBT'', ``ANN'', and ``SpecTf'' are binary cloud masks produced by each respective model. \textbf{b} displays the same products for the held-out EMIT scene \texttt{emit20230425t082443}. \textbf{c}, Cloud detection ROC curves on the held-out dataset of the different models. \textbf{d}, Enlarged view of the upper left section of the ROC curve for clearer distinction between models.}
    \label{fig:result-masks}
\end{figure}

Quantitative differences between ML models are minute relative to the difference between ML models and the baseline. The GBT insignificantly outperforms SpecTf in ROC AUC by 0.005, and the ANN insignificantly outperforms SpecTf in $F_{1}$-score by 0.004. The nearly overlapping ROC curves of the ML models (Fig.~\ref{fig:result-masks}c,d) further emphasizes the similarities. Notably, however, the models differ dramatically in the number of learned parameters required, with SpecTf needing only 20,000 learned parameters (two orders of magnitude fewer than the ANN), less than half of even the number of feature splits (decision nodes) in the GBT.

Large qualitative differences in model performance reveal additional important distinguishing characteristics when the models are applied to complete EMIT scenes. As demonstrated quantitatively, all three ML models outperform the baseline;  Figure~\ref{fig:result-masks}a, shows how the GBT, ANN, and SpecTf masks all successfully detect a large cloud bank at the center of the scene that the baseline mask missed. Differences between the ML masks are more distinct in the cloud masks shown in Figure~\ref{fig:result-masks}b. This challenging scene contains visually clear cirrus clouds, highlighted in yellow-green in the false color image. All ML models are able to detect and mask these clouds where the baseline fails. However, the GBT and ANN masks also contain notable vertical streaking effects, a common failure mode of ML models applied to imaging spectroscopy data due to subtle wavelength differences in the crosstrack dimension of the sensor. The vertical streaks include both false positive and false negatives in different portions of the scene, which stand out due to the lack of spatial cohesiveness common to true clouds. Models that lean strongly on very specific thresholds are susceptible to this failure mode, given the slight differences in incident irradiation at the sensor as a function of wavelength variation in the crosstrack. The demonstrated brittleness of the reference ML models is a core challenge to large-scale deployment of ML models in spectroscopy. Despite also being a single spectrum model (no spatial context is utilized), the SpecTf does not exhibit this vertical streaking behavior, instead producing a clean, spatially coherent cloud mask that accurately detects the cirrus clouds. While all models are bound to produce some false positive detections in challenging scenes, the failure mode of SpecTf is less susceptible to these unwanted mask artifacts. A relevant factor may be the binary threshold of each model---while the best-performing probability thresholds for the GBT and ANN are 0.96 and 0.98, the SpecTf threshold is 0.52, which indicates that the model's posterior probability is better calibrated. It also suggests that a sequential representation of the spectra is more robust to variations between adjacent pixels, compared to treating each wavelength as an independent input feature.


\subsection{Interpreting model predictions with attention weights} \label{sec:results-interp}


SpecTf and its transformer architecture is inherently interpretable, providing a transparent view of the rationale behind individual predictions. Most ML models, including our reference cases, require a \textit{post hoc} method to derive feature importance, such as SHAP (a game-theoretic approach) \citep{lundberg2017shap}, Integrated Gradients (a gradient approach) \citep{qi2019ig}, and other methods that attempt to interpret the learned weights of a deep learning or tree-based model. Instead, we are able to directly interpret the weights of the self-attention layer to scale the contribution of a wavelength to the model's output. These weights are learned and calculated by evaluating the appropriate value of wavelengths relative to each other. Summing the two dimensional relational weights into one of its axes, we can visualize the relative contribution of different wavelengths. Spectroscopically, this gives insight into which absorption features dictate a particular classification decision.


\begin{figure}[t]
    \centering
    \includegraphics[width=0.7\linewidth]{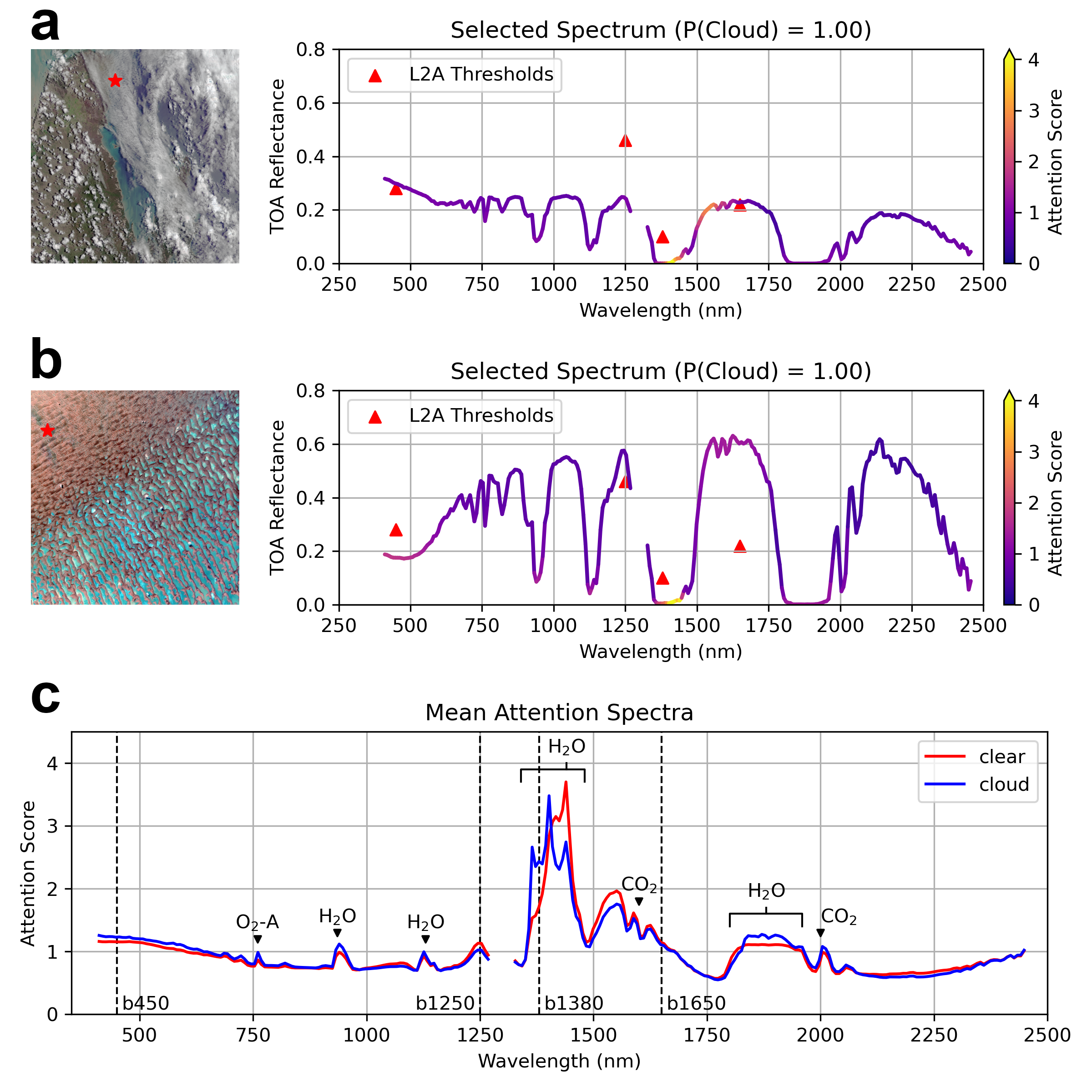}
    \caption{\textbf{a}, Interpretation of an individual spectrum in \texttt{emit20240302t005829} incorrectly classified by the baseline model and correctly identified as a cloud by SpecTf. A higher attention score (brighter color) reflects the wavelength's greater contribution to the prediction. \textbf{b}, Interpretation of a spectrum in \texttt{emit20230425t082443} incorrectly missed by the baseline model and correctly identified as a cloud by SpecTf. The thresholds used by the baseline model to detect clouds are marked with red triangles in \textbf{a} and \textbf{b}.
    \textbf{c}, Average attention scores across all pixels labeled as ``clear'' and ``cloud'' in red and blue, respectively. Atmospheric absorption features of O$_2$, CO$_2$, and H$_2$O identified by the model are annotated for reference. The four wavelengths used by the baseline (450, 1250, 1380, 1650 nm) are marked with dashed lines.}
    \label{fig:result-interp}
\end{figure}

Figure~\ref{fig:result-interp}a shows an example of SpecTf prediction interpretation. The baseline model did not identify this spectrum as a cloud because it fell below the threshold at 1250 nm and 1380 nm (denoted with red triangles), whereas SpecTf correctly made a cloud prediction with 100\% confidence. The attention weights, represented as a colormap over the spectrum, highlight the 1409 nm and 1551 nm bands as important contributors to this prediction. Figure~\ref{fig:result-interp}b shows a second example. This spectrum fell below the thresholds at 450 nm and 1380 nm, leading to a misclassification as a clear pixel, whereas SpecTf correctly made a cloud detection with 100\% confidence. In addition to the emphasis of 1417 nm, attention weights are elevated around 450 nm and 1625 nm.

Finally, by averaging the attention weights of predictions over the entire labeled dataset, we can interpret which wavelengths SpecTf found most relevant for the cloud detection task. Figure~\ref{fig:result-interp}c shows these ``mean attention spectra'' for clear and cloud spectra, separately. This figure clearly shows that SpecTf has learned, and is paying attention to, known atmospheric gas absorption features. Even when capturing sharp absorption features, the attention scores highlight a relatively smoothly varying importance, indicating the model is utilizing the spectral context of each absorption. This contrasts classical behavior of ML models in spectroscopy that lean on individual wavelengths independently. Peaks in attention score are located at known absorption wavelengths of H$_2$O, O$_2$-A, and CO$_2$, with the most significant contributions occurring between $1300$ nm and $1500$ nm, a significant water vapor absorption feature.  These wavelengths are physically significant because sunlight in these channels is completely absorbed by water vapor in the lower troposphere. Thus, their radiance is near zero in the absence of clouds.  When clouds are present, suspended liquid and/or ice particles scatter light back to the sensor before it reaches the absorbing atmospheric layer, increasing the signal in these wavelengths \cite{gao1993cirrus}.  Consequently it makes sense that the edges and floor of this feature could be used to recognize clouds. Interestingly, the oxygen A band at 760 nm, which has been used previously as a surface elevation cue for cloud detection in sounding spectrometers \cite{taylor2011comparison}, receives far less attention.  It is possible that non-cloud sources of A band variance such as the ground elevation and/or aerosol scattering in the near infrared make this channel less effective than the longer wavelengths available to EMIT.  

\subsection{Generalizing to data from different instruments and platforms}

\begin{figure}[t]
    \centering
    \includegraphics[width=0.7\linewidth]{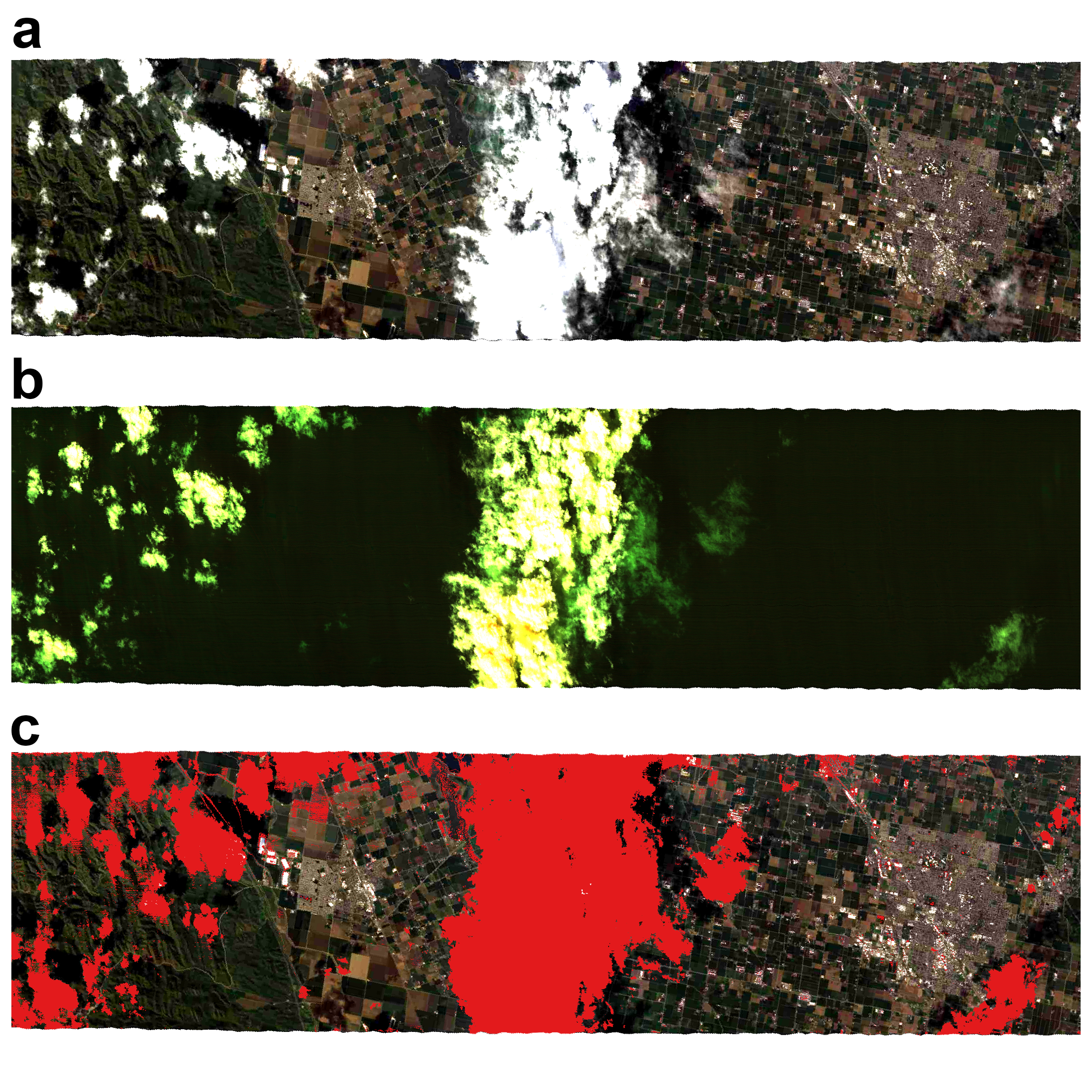}
    \caption{Results of SpecTf trained on EMIT data and applied to a flightline observed by the AVIRIS-NG instrument from the ER-2 airborne platform as part of the GOES-R checkout campaign. \textbf{a}, RGB quicklook of a section of the \texttt{ang20170321t213141} flightline. \textbf{b}, false color visualization of the 1380, 1420, and 1890 nm wavelengths for easier cloud identification. \textbf{c}, cloud mask produced by the SpecTf model trained on EMIT data in red, overlaid on RGB. All clouds present are detected, but some false positive detections are present around anthropogenic regions.}
    \label{fig:result-avng}
\end{figure}

SpecTf is a sequence model, where the input length can be variable. SpecTf learns to embed each input spectrum's representation from each band's wavelength and TOA reflectance. These two elements of the model architecture allows us to apply SpecTf, trained only on EMIT spectra, on data observed by a different instrument with a different set of wavelengths. This is possible without any retraining or modifications to the model for a reasonably similar spectral range; it is impossible to perform with the GBT and ANN models, which expect independent input features of constant length at fixed wavelengths.

Figure~\ref{fig:result-avng} shows the result of applying the SpecTf, trained only on EMIT data, on a scene collected by AVIRIS-NG, an airborne imaging spectroscopy instrument. While EMIT data has 285 bands at 7.5 nm resolution from 380 nm to 2500 nm, this scene, observed from the ER-2 airborne platform, has 425 bands at 5 nm resolution of the same spectral range. Despite 140 more input features, a different spectral resolution, and many other confounding factors borne of a different instrument on a different platform (e.g. instrument sensitivity, spatial resolution), SpecTf is able to produce a reasonable cloud mask for this scene (Fig.~\ref{fig:result-avng}c). It successfully masks all clouds present, including some translucent clouds only faintly visible in the false color image (Fig.~\ref{fig:result-avng}b). Minor false detections are present over agricultural fields and urban areas, where diverse materials present a challenge for spectroscopy tasks. Refining the model by training it on more agricultural and urban scenes (underrepresented in the original training dataset), and on data from multiple instruments (either in the initial training or after the fact), would almost certainly improve the performance. However, the zero-shot performance, combined with model interpretation, is a strong indicator that SpecTf was able to learn a physically meaningful representation of the reflectance spectrum as a whole, not just independent features from each wavelength.



\section{Discussion}

In this work, we propose the Spectroscopic Transformer as an accurate, interpretable, and instrument-agnostic model to produce cloud masks for the EMIT imaging spectroscopy mission. Our quantitative evaluation demonstrated SpecTf's significant cloud detection performance improvement over the currently produced cloud mask, as well as improved model efficiency over other reference-case ML models. Qualitative assessment of produced cloud masks demonstrated SpecTf's robustness to sensor noise and superior spatial coherence relative to ML approaches despite it being a pixelwise model. Model interpretation with attention scores clearly showed that SpecTf has learned meaningful physical phenomena for the cloud detection task. Finally, SpecTf's architecture learns properties of wavelengths in the continuous electromagnetic spectrum, rather than patterns of a particular channelization. This allows it to generalize beyond the instrument it was trained on, performing well on data collected by different instruments on different platforms. In total, the Spectroscopic Transformer demonstrates several attractive capabilities that prove it to be a powerful methodology for data-driven imaging spectroscopy tasks.

While many deep learning based methodologies for remote sensing cloud detection exist, most models are based on Convolutional Neural Network (CNN, e.g. U-Net) or Vision Transformer (ViT) architectures that rely on spatial features for detection, intended for multi-spectral data such as Sentinel 2 (13 bands) or Landsat 8 (11 bands) \citep{pasquarella2023csplus, dronner2018fast, zhang2022cloudvit}. CNNs are infeasible for hyperspectral images without an external or embedded dimensionality reduction, as convolutional filters scale poorly for hundreds of input channels. While existing methods such as SpectralFormer \citep{hong2021spectralformer} and Spatial-Spectral Transformers \citep{he2021spatial} have applied transformers and ViTs to hyperspectral images for land type classification, these methods continue to rely on image patches to learn spatial features. These prior works extend computer vision techniques for RGB images to imaging spectroscopy, often neglecting the value of the the spectroscopy data in hand. For tasks where sufficient relevant information is contained within the spectra, learning spatial features (and therefore spatial biases) becomes a liability; models cannot ignore spatial features when they become uninformative or irrelevant. This is especially the case for cloud detection, where clouds present with incredibly diverse shapes and are often amorphous without clear edges. A pixelwise model that makes an independent determination with each spectrum will always be superior to a spatial model if spatial features are unnecessary to compensate for the lack of spectral information. We demonstrated that water absorption features and surrounding spectral context are more than sufficient for SpecTf to detect clouds pixel by pixel.


Physically meaningful interpretability is a valuable - and perhaps even necessary - capability for deep learning models performing scientific tasks. Model interpretation with an inherent mechanism of the model architecture is even more valuable than model interpretation via \textit{post hoc} evaluation of model weights. Visualizing the attention weights of the SpecTf shows that the model has learned real atmospheric absorption features to perform the cloud detection task. The attention weights shown in the mean attention spectra demonstrate that the model uses the spectrum as a whole, and utilizes spectral continuity; this is highly unlikely to be realized with GBT and ANN models that treat each wavelength as independent, unrelated input features. Model interpretation makes it clear that SpecTf correctly treats the input sequence as a discrete sampling of an underlying continuous spectrum. This, in turn, enables SpecTf to generalize to input data from different instruments. Instrument-agnostic algorithms are growing in necessity as the number of current and future imaging spectroscopy instruments increase (e.g. AVIRIS-3, SBG VSWIR, Carbon Mapper~/~Tanager, CHIME, Carbon-I, etc.). In future work, SpecTf could be trained on data from multiple such instruments, establishing a unified pipeline for cloud masking and other spectral tasks. Some such tasks may require modifying the current input vector, which assigns each spectral channel a single wavelength identity independent of its true spectral response function.   In reality, two different instruments' response functions may differ dramatically even if the channel center wavelengths are the same, which would influence the measured values of the radiance spectrum. However, we see no reason that a future iteration of SpecTf could not learn to utilize the spectral response function of multiple instruments during training.

SpecTf still has limitations common with many data-driven methods. While SpecTf did significantly reduce the number of false positive detections in high-altitude snow regions compared to the baseline (e.g. Himalayas), some false positive detections remain. Performance in these regions could be further improved by sampling more training examples or by including geographic features. As an ML model, SpecTf is limited by the comprehensiveness of its training dataset, and ``unknown unknown'' spectra outside of its dataset will risk unexpected behavior that is difficult to constrain. The current version of SpecTf also does not distinguish between opaque clouds (e.g. cumulus) and translucent clouds (e.g. cirrus), in large part due to the difficulty of consistently and accurately labeling and annotating both classes simultaneously. Supervised machine learning models are limited to tasks with clear ground truth data; while a model that produces a cloud transparency percentage map would be useful, producing the ground-truth dataset for such as task would require vast simulations, and then be subject to the efficacy of the simulation framework. Finally, model interpretation with attention scores is limited to feature importance, and does not indicate directionality (whether an input feature contributed positively or negatively to the predicted class).

While SpecTf was originally developed for the cloud masking task, it was designed with other spectral tasks in mind, such as retrievals of atmospheric and surface properties. This simple model architecture can be easily scaled and modified to accommodate more complex tasks or multiple input spectra. For example, the model could process both an input spectrum and a target absorption spectrum, or process two spectra observed at different time points. We plan to extend SpecTf to several such tasks in the future. In this study, we also limited the training dataset to manually annotated data from EMIT; simulation-based data likely have a meaningful role to play in future iterations of Spectroscopic Transformers.

\section{Methods}

\subsection{Data preparation and labeling}

EMIT scenes were selected and human-annotated for model training and evaluation. 221 scenes were randomly selected after stratifying by sun angle, L2A cloud cover percentage, and water vapor content to ensure a diversity of clear and cloudy scenes. After a period of development, a preliminary model was deployed on all scenes observed during the month of March 2024, from which another 313 scenes were selected to correct false predictions.

The first batch of EMIT scenes were labeled with the image semantic segmentation feature of Labelbox, a commercial data annotation platform. Scenes were sparsely labeled with pixelwise masks for the ``clear'', ``cloud'', ``cloud shadow'', and ``optically clear cirrus'' classes. Afterwards, the ``clear'' and ``cloud shadow'' classes were combined, and the ``cloud'' and ``optically clear cirrus'' classes were combined, due to ambiguity and uncertainty during labeling. Cloud shadows were frequently covered by a thin layer of translucent clouds, and the boundary between opaque and clear clouds was unquantifiable, leading to confusion in determining the ground truth. Only pixels that could be confidently determined by labelers as cloud-covered or clear were labeled, and unlabeled pixels were not used.

The second batch of EMIT scenes were labeled with the polygon annotation feature of the Multi-Mission Geographic Information System (MMGIS) \citep{calef2024mmgis}. These annotations focused on correcting false positive or false negative detections by the model. Systemic false detections at this point included false positives over oceans, snowy mountains, and coastlines. Only the ``cloud'' and ``clear'' classes were labeled, and vector shape annotations were rasterized into pixelwise label masks to match the Labelbox masks.

After collecting the non-orthorectified EMIT L1B radiance rasters and the label mask rasters, we calculated Top-of-Atmosphere (TOA) reflectance rasters with the observational geometry products. This provided a degree of input data normalization for the machine learning models, as TOA reflectance values are between 0 and 1 (aside from some artifacts that may slightly exceed 1). Finally, we randomly sampled up to 10,000 pixels from each class in each scene. This ensured that completely cloudy or completely clear scenes (up to 2 million pixels each) did not dominate other scenes with smaller clouds or clear regions. A balance of diverse examples is critical to a globally generalizable model.

3,575,442 pixels were sampled in total: 1,642,181 ``cloud'' pixels and 1,933,261 ``clear'' pixels. This dataset was split into training and validation datasets, stratified to ensure that no scene contributed pixels to both the training and validation datasets. Of the 534 scenes, 465 scenes comprised the training set, and 69 scenes comprised the validation set: 3,078,931 pixels and 496,511 pixels, respectively.

For each input EMIT spectrum, we dropped bands between 380-400 nm and 2450-2500 nm due to potential artifacts from the de-striping algorithm used by EMIT. We also dropped bands between 1275-1320 nm to avoid the order sorting filter boundary. In total, 17 bands were removed, reducing the number of bands from 285 to 268. When applying the model to spectra from AVIRIS-NG, all 425 bands were included in the input.

\subsection{L2A baseline and reference model architectures}

The EMIT L2A cloud filter is a combination of a trivariate band threshold for opaque clouds and a single band threshold for cirrus clouds \citep{thompson2024atbd}. This filter is defined in Equation~\ref{eq:l2a}, where ``b450'' represents the TOA reflectance at the 450 nm wavelength, etc. A slightly simplified version of this filter is actively run in-orbit on EMIT's Field Programmable Gate Array (FPGA) to maximize the EMIT data yield with constrained downlink, and the full version currently runs on the ground to produce the L2A cloud mask product.

\begin{equation} \label{eq:l2a}
    ((\text{b450} > 0.28) \land (\text{b1250} > 0.46) \land (\text{b1650} > 0.22)) \lor (\text{b1380} > 0.1)
\end{equation}

Two reference classification models were developed to quantitatively and qualitatively evaluate the performance of the SpecTf model. First, the Gradient Boosted Tree (GBT) classification model was implemented using the Python XGBoost library \citep{chen2016xgboost}. GBTs are known for being capable and efficient models that offer an alternative mechanism than learning via latent representations, which is useful as a competitive benchmark for deep learning methods on this classification task. After a hyperparameter grid search, an architecture with 300 estimators and maximum depth of 6 was chosen as the best-performing model by the $F_1$-score metric on the validation dataset; all other hyperparameters were found to be negligible. The model was trained with default optimization parameters.


\begin{table}[t]
\centering
    \begin{tabular}{l|r|r|r}
    Layer Type & Input Dims. & Output Dims. & Parameters \\
    \hline\hline
        \textbf{Residual Connection} & & & \\
        Linear & 268 & 1400 & 376,600 \\
        LayerNorm & 1400 & 1400 & \\
    \hline
        \textbf{Main Branch} & & & \\
        Linear & 268 & 1400 & 376,600 \\
        LayerNorm & 1400 & 1400 & \\
        GeLU Activation& 1400 & 1400 & \\
        Dropout (0.2) & 1400 & 1400 & \\
        Linear & 1400 & 1400 & 1,961,400 \\
        LayerNorm & 1400 & 1400 & \\
        GeLU Activation& 1400 & 1400 & \\
        Dropout (0.2) & 1400 & 1400 & \\
        \textit{Residual Addition} & 1400 & 1400 & \\
        GeLU Activation& 1400 & 1400 & \\
        Linear & 1400 & 2 & 2,802 \\
        Softmax & 2 & 2 &
    \end{tabular}
\caption{ANN model architecture. This model has a total of $2 \times 10^6$ learned parameters. Dropout layers are only applied during model training. \textit{Residual Addition} refers to the output of the \textbf{Residual Connection} being added with the output of the previous layer to form a skip connection.}
\label{tab:ann_arch}
\end{table}

Second, the Artificial Neural Network (ANN) model was implemented using the Pytorch library \citep{paszke2019pytorch}. The ANN model was implemented as a residual network \citep{he2016deep} where the outputs of the feed forward dense layers were added together with inputs (known as skip or highway connections), which demonstrated slightly better performance than feed forward dense layers alone. A hyperparameter grid search over the width, depth, batch size, nonlinearity, and normalization method found that a single hidden layer model with a width of 1400 projection dimensions to perform best on the $F_1$-score metric on the validation dataset. Table~\ref{tab:ann_arch} describes the architecture of the ANN model in full. The model was trained to convergence with the Schedule-Free AdamW optimizer \citep{defazio2024road} for 30 epochs with a batch size of 1024 and learning rate of $1 \times 10^{-5}$.

\subsection{Spectroscopic Transformer model architecture}

\begin{figure}[th]
    \centering
    \includegraphics[width=0.5\linewidth]{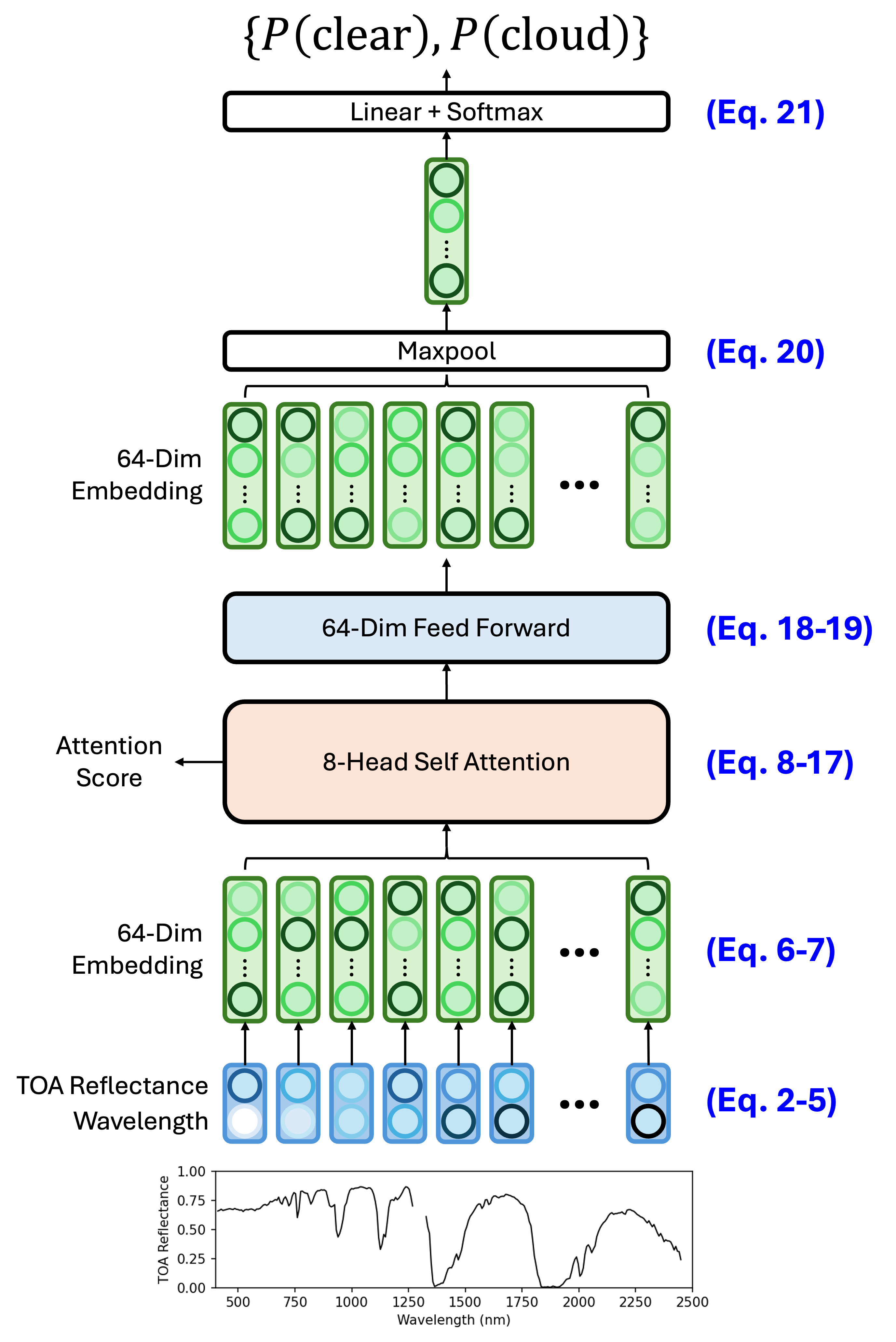}
    \caption{The Spectroscopic Transformer model architecture annotated with equation references.}
    \label{fig:arch-eq}
\end{figure}

The Spectroscopic Transformer is heavily based on the encoder module of the transformer architecture by \cite{vaswani2017transformer} and is implemented in Pytorch \citep{paszke2019pytorch}. Major differences include a different positional encoding, lack of residual connections, and the addition of the maxpool classification head. We now describe this architecture in detail, in conjunction with an annotated architecture diagram (Fig. \ref{fig:arch-eq}).

First, we define our input: the top-of-atmosphere (TOA) reflectance spectrum $\mathbf{s}$ (Eq.~\ref{eq:s}) and band wavelength centers $\mathbf{b}$ (Eq.~\ref{eq:b}). For EMIT, these sequences are $n=268$ long. While $\mathbf{s}$ is already mostly normalized in the range $[0, 1]$, $\mathbf{b}$ needs to be roughly centered and scaled to a mean of $0$ and standard deviation of $1$ for improved model training and convergence (Eq.~\ref{eq:b_pre}). Finally, the input sequence $X^{(0)}$ is defined as a sequence of reflectance and wavelength pairs $\{s_i, b'_i\}$ (Eq.~\ref{eq:input}). The model will determine if a pixel is clear or cloudy based on this sequence of pairs alone. Note that we are omitting the batch dimension, commonly used to compute multiple inputs in parallel, for simplicity.

\begin{align}
    \mathbf{s} &= \{s_1, s_2, \ldots, s_n\} \in \mathbb{R}^n && \text{TOA Refl. Spectrum} \label{eq:s}\\
    \mathbf{b} &= \{b_1, b_2, \ldots, b_n\} \in \mathbb{R}^n && \text{Band wavelengths} \label{eq:b}\\
    \mathbf{b'} &= (\mathbf{b} - 1440) / 600 && \text{Simple centering} \label{eq:b_pre}\\
    X^{(0)} &= \left\{ \{s_i, b'_i\} \mid i \in [1..n] \right\} \in \mathbb{R}^{n \times 2} && \text{Input vector} \label{eq:input}
\end{align}

The first layer of SpecTf projects each $\{s_i, b'_i\}$ pair into a higher dimensional embedding. A fully-connected layer learns to project each item into $d_{\text{model}}=64$ dimensions, after which a tanh activation normalizes the embedded values to prevent exploding gradients during training (Eq.~\ref{eq:proj}). This projection is driven by mechanical necessity; low dimensionality of each item's representation directly limits the model's capacity to learn, especially during the dot product calculations during attention. An unnecessarily high dimensional representation, however, will significantly increase computational complexity with insignificant improvements to task performance and risk overfitting. After the embedding projection, layer normalization helps the model learn more effectively by standardizing the inputs within each layer, ensuring that the values remains in a consistent and manageable range throughout the network (Eq.~\ref{eq:layernorm}).

\begin{align}
    X^{(1)} &= \{\tanh(\text{Linear}_{2\rightarrow64}(X^{(0)}_i)) \mid i \in [1..n] \} \in \mathbb{R}^{n \times 64} && \text{Embedding projection} \label{eq:proj}\\
    X^{(2)} &= \text{LayerNorm}(X^{(1)}) && \text{Normalization} \label{eq:layernorm}
\end{align}

We now describe the self-attention layer. First, the embeddings are linearly projected (without a nonlinear activation function) into the ``Query'' $Q$, ``Key'' $K$, and ``Value'' $V$ representations (Eq.~\ref{eq:q_def},\ref{eq:k_def},\ref{eq:v_def}). We will discuss the dimensionality of these representations ($d_q$, $d_k$, and $d_v$) when describing the multi-head extension later. Each of these learned representations will serve a distinct role in the attention mechanism, although their inputs, $X^{(2)}$, are the same.

\begin{align}
    Q &= \{ \text{Linear}_{64\rightarrow d_q}(X^{(2)}_i) \mid i \in [1..n] \} \in \mathbb{R}^{n \times d_q} && \text{Query linear proj.} \label{eq:q_def}\\*
    K &= \{ \text{Linear}_{64\rightarrow d_k}(X^{(2)}_i) \mid i \in [1..n] \} \in \mathbb{R}^{n \times d_k} && \text{Key linear proj.} \label{eq:k_def}\\*
    V &= \{ \text{Linear}_{64\rightarrow d_v}(X^{(2)}_i) \mid i \in [1..n] \} \in \mathbb{R}^{n \times d_v} && \text{Value linear proj.} \label{eq:v_def}
\end{align}

We now step through the attention layer. First, for each item $i$ in the sequence $[1..n]$, we take the dot product between its Query $Q_i$ and the Key $K_{j\in[1..n]}$ of every item $j$ in the sequence (including itself). The scalar product $W_{ij}$ of each Query-Key dot product quantifies the relevancy of $j$ to $i$ for the task at hand (Eq.~\ref{eq:attn_qk}). In the context of SpecTf, this operation reveals the interdependencies between wavelengths that are most informative for cloud detection. Comparing wavelengths in this manner is what allows SpecTf to efficiently and meaningfully learn spectral tasks. Collecting all dot products $W_i = \{W_{ij} \mid j \in [1..n]\}$, scaling them by the representation dimension $d_k$ (for gradient stability), then taking the softmax (such that all $W_{ij}$ for a given $i$ sum to $1$) results in the attention weights $W_i$ (Eq.~\ref{eq:attn_w}).

This quantification of the relative importance of items in the sequence is used to weight the information that percolates to the rest of the network. The Value representation $V_j$ represents the information contained in every item $j\in[1..n]$, and the weighted sum of these Values by $W_i$ becomes the output $Y_i$ for the given item $i$ (Eq. \ref{eq:attn_out}). In terms of SpecTf, the attention weights scale the output such that signals from relevant wavelengths are included, and signals from irrelevant, uncorrelated, or confounding wavelengths are discarded. An attention weight $W_{ij}=0$ indicates that wavelength $j$ is not relevant to wavelength $i$, and that its information $V_j$ should not be included in the output of $i$.

Packing all queries, keys, and values into $Q$, $K$, and $V$ allows us to calculate outputs for all Queries simultaneously with computationally efficient matrix multiplication, producing the output $Y$ (Eq. \ref{eq:attn}). Note that the attention mechanism itself is only a series of matrix multiplications without any learned parameters. Instead, it relies on the linear projections that produced the Queries, Keys, and Values to learn useful representations that effectively serve their respective roles in the attention mechanism.

\begin{align}
    W_{ij} &= Q_iK_j^T \in \mathbb{R} && \text{Query-key dot product} \label{eq:attn_qk}\\
    W_i &= \text{softmax}\left(\frac{Q_i K^T}{\sqrt{d_k}}\right) \in \mathbb{R}^{n} && \text{Attention weights of item } i \label{eq:attn_w}\\
    Y_i &= W_i V \in \mathbb{R}^{d_v} && \text{Attention outputs of item } i \label{eq:attn_out}\\
    Y &= \text{Attention}(Q, K, V) && \text{Complete attention definition} \label{eq:attn}\\*
      &= \text{softmax} \left(\frac{QK^T}{\sqrt{d_k}}\right)V \in \mathbb{R}^{n \times d_v}  \nonumber
\end{align}

So far, we have described Scaled Dot-Product Attention in context of the spectral task. As an extension, we now describe the actual Multi-Head Self-Attention layer implemented in SpecTf, which allows the model to perform multiple attention functions in parallel without increased computational complexity. The implementation is straightforward: for each of the $h=8$ attention heads, we learn a different set of Query, Key, and Value representations of dimensions $d_q = d_k = d_v = d_{\text{model}} / h = 64 / 8 = 8$ (Eq.~\ref{eq:head_def}). Note that the number of heads and the dimensionality of the representations is limited by $d_{\text{model}}$, motivating the earlier embedding projection to 64 dimensions in Equation~\ref{eq:proj}. The output $Y^i \in \mathbb{R}^{8\times n}$ of each attention head is then concatenated, before a fully-connected linear produces the final output $X^{(3)}$ (Eq.~\ref{eq:mh_attn}). In terms of the SpecTf, this allows the model to learn different types of relationships between wavelengths in parallel.

\begin{align}
    \text{head}^i &= \text{Attention}(Q^i, K^i, V^i) \mid i \in [1..h] && \text{Attention head def.} \label{eq:head_def}\\*
                  &\text{where } Q^i \in \mathbb{R}^{n \times (d_\text{model}/h)} \nonumber \\
    X^{(3)} &= \text{Linear}_{64\rightarrow64}(\text{Concat}(\text{head}^1, \ldots, \text{head}^8)) && \text{Multi-Head Self-Attention} \label{eq:mh_attn} \\*
      &\in \mathbb{R}^{n\times64} \nonumber\\
    X^{(4)} &= \text{LayerNorm}(X^{(3)}) && \text{Normalization}
\end{align}

In the original transformer architecture, there is a residual skip connection which adds the input sequence $X^{(2)}$ to the output of the Multi-Head Self-Attention layer $X^{(3)}$. This connection is necessary in large language models or vision transformers where many stacked encoder modules cause vanishing gradients during model training and backpropagation. This residual skip connection is unnecessary for SpecTf, which only uses a single attention layer. In fact, we determined that it sometimes caused the model to ignore the attention layer during training, resulting in worse performance and interpretability.

The rest of the model architecture is aimed at extracting classification information from the attention output and predicting a class. A feed forward module first applies two more fully-connected layers to each item in the sequence to extract signals from the relevant information identified by the attention module (Eq.~\ref{eq:ff_1},\ref{eq:ff_2}). Penultimately, we aggregate the signals contained in the entire sequence by maxpooling across $n$; that is, taking the maximum value of each representation dimension across the length of the sequence (Eq. \ref{eq:maxpool}). This produces a single 64-dimension vector that is finally classified by a single fully-connected layer and the softmax function to produce two probabilities for the ``clear'' and ``cloud'' classes that sum to $1$ (Eq. \ref{eq:class_head}).

\begin{align}
    X^{(5)} &= \{\text{GeLU}(\text{Linear}_{64\rightarrow64}(X^{(4)}_i)) \mid i \in [1..n]\} \in \mathbb{R}^{n\times64} && \text{FF first layer} \label{eq:ff_1}\\
    X^{(6)} &= \{\text{Linear}_{64\rightarrow64}(X^{(5)}_i) \mid i \in [1..n]\} \in \mathbb{R}^{n\times64} && \text{FF second layer} \label{eq:ff_2}\\
    \mathbf{x}^{(7)} &= \max_{i \in [1..n]} X^{(6)}_i \in \mathbb{R}^{64} && \text{Maxpool} \label{eq:maxpool}\\
    P(\text{clear} | X)&, P(\text{cloud} | X) = \text{Softmax}\left(\text{Linear}_{64\rightarrow 2}(\mathbf{x}^{(7)})\right) \in \mathbb{R}^2 && \text{Classification} \label{eq:class_head}
\end{align}

This classification head is different from the ``class token'' prepended to the input sequence by ViT \citep{dosovitskiy2021vit} and BERT \citep{kenton2019bert} to accumulate classification information. We did not consider this methodology because SpecTf only uses a single encoder layer, compared to ViT and BERT's 12 encoder layers. A class token simply would not have enough opportunities to aggregate classification information from a single attention layer. Therefore, we only considered classification heads that would process the output sequence as a whole. Not aggregating and inputting the flattened sequence as $n\times64$ independent features into a feed-forward network would fix the length of the sequence, eliminating the capability to generalize to other spectra with different lengths. Understanding that it was necessary to aggregate across the sequence, we deemed likely that the majority of wavelengths in the sequence would not contain any signals relevant to cloud detection. Aggregating with a maximum across the sequence (instead of an average) allows a single wavelength with high signal to override many other wavelengths with low signals. We also confirmed experimentally that flattening and averaging classification heads performed worse than the maxpooling classification head.

Architectural decisions and model hyperparameters were determined through model architecture and hyperparameter optimization using the Bayesian search feature of Sweeps by Weights \& Biases, a commercial AI development platform. Over the course of development, nearly 2,000 models were trained while varying the model architecture (e.g. number of encoder layers, positional encoding, aggregation functions, etc.) and hyperparameters (e.g. $d_{\text{model}}$, $h$, feed forward layer size, etc.). Earlier versions of the model also included a 1-dimensional convolution of a 7-dimensional kernel over the input spectrum to group adjacent wavelengths (in place of Eq. \ref{eq:proj}); however, we found that this did not improve performance, and negatively impacted the clarity of model interpretation.

The final reported model was trained to convergence for 30 epochs of 3 million spectra, which took 9.5 hours on a Nvidia A100 GPU. The model was trained with the Schedule-Free AdamW optimizer \citep{defazio2024road} with a learning rate of $1 \times 10^{-4}$, a batch size of 256, and dropout rate of 0.1 where applicable. 

\subsection{Attention weight spectrum}

As described in Section \ref{sec:results-interp}, the self-attention layer produces attention weights that can be interpreted as the importance of each wavelength given an input spectrum. Recall from Equations \ref{eq:attn_qk}-\ref{eq:attn_out} that the attention mechanism uses the attention weight $w_{ij}$ to quantify the relevancy of wavelength $j$ to wavelength $i$ and weight how much of its Value $V_j$ is included in the output $Y_i$ (Eq. \ref{eq:attn_w2}). While the attention weights $W_{i, j\in[1..n]}$ for each Query $Q_i$ sums to 1 (for the weighted sum of $V_i$), the attention weights $W_{i\in[1..n],j}$ for each Key $K_j$ does not---in fact, their sum represents how highly weighted each Key was by every Query of the sequence. We define this sum of each Key's attention weights as the attention spectrum (Eq. \ref{eq:attn_score}). Wavelengths that contain information that is highly relevant to the rest of the input spectrum receives high attention weights, and therefore sums to a high attention value.

\begin{align}
    &W \in \mathbb{R}^{n \times n} \text{ where } W_{ij} = Q_i K^T_j && \text{Attention Weight Matrix} \label{eq:attn_w2} \\*
    &\left\{ \sum_{i=1}^{n} W_{ij} \mid j \in [1..n] \right\} \in \mathbb{R}^n && \text{Attention Weight Spectrum} \label{eq:attn_score}
\end{align}

While attention weights are useful as an inherent feature importance product of the model, it does not provide directional (positive or negative) explanations for the output class as other \textit{post hoc} methods can (e.g. SHAP \citep{lundberg2017shap}). Additional extensions to the attention mechanism such as AGrad \citep{liu2021exploring} could provide this capability.

\subsection{Generalization to AVIRIS-NG spectra}

Applying the SpecTf model trained on EMIT data to detect clouds in AVIRIS-NG data requires no architectural changes to the model. While AVIRIS-NG data has 425 bands at 5 nm (compared to EMIT's 285 bands at 7.5 nm), the only variables that change are in the wavelengths of the input sequence $\mathbf{b}$ (which were constant for all EMIT data), and $n$ (the length of the sequence). First, because $\mathbf{b}$ is embedded into a higher dimensional representation, the model is able to smoothly embed wavelengths that fall in between EMIT wavelengths it was trained on. Second, because SpecTf is a sequential model, $n$ is a free variable that can change during training and inference.

Future work to train a model on data from multiple instruments may consider including other information to improve inter-instrument generalization. For example, the full width at half maximum of the instrument may be included to consider differences in spectral resolution, or some normalization for the atmospheric thickness may account for TOA Reflectance difference between airborne and spaceborne platforms. Training on sequences of different lengths will also require implementation of existing padding and masking mechanisms for the matrix multiplication operations in the model.

\section{Data availability}

EMIT L1B at-sensor calibrated radiance and geolocation data is available at: \url{https://lpdaac.usgs.gov/products/emitl1bradv001/}. Annotated data for training and evaluation of the models in this work, as well as complete cloud mask results on the validation dataset, are available at: \url{https://doi.org/10.5281/zenodo.14607937}

\section{Code availability}

All EMIT data product algorithms and workflows are available at: \url{https://github.com/emit-sds}. Source codes for this work, including trained model weights, are available at: \url{https://github.com/emit-sds/SpecTf}. 

\section{Author contributions}
J.H.L and P.G.B conceived the study. J.H.L, M.K., and P.G.B. designed the model architecture and training/evaluation system. M.K. and J.H.L. labeled training data, with review from P.G.B.  J.H.L. and M.K. wrote and trained the model.  P.G.B. and D.R.T. provided spectroscopic interpretation.  All authors contributed critically to the manuscript drafts and approved the publication.

\backmatter

\bmhead{Acknowledgements}

The EMIT online mapping tool was developed by the JPL MMGIS team. The High Performance Computing resources used in this investigation were provided by funding from the JPL Information and Technology Solutions Directorate. The research was carried out at the Jet Propulsion Laboratory, California Institute of Technology, under a contract with the National Aeronautics and Space Administration (80NM0018D0004). © 2024. California Institute of Technology. Government sponsorship acknowledged.





\bibliography{sn-bibliography}

\begin{thebibliography}{10}
\expandafter\ifx\csname url\endcsname\relax
  \def\url#1{\burl{#1}}\fi
\expandafter\ifx\csname urlprefix\endcsname\relax\def\urlprefix{URL }\fi
\providecommand{\bibinfo}[2]{#2}
\providecommand{\eprint}[2][]{\url{#2}}
\providecommand{\doi}[1]{\url{https://doi.org/#1}}
\bibcommenthead

\bibitem{green2020emit}
\bibinfo{author}{Green, R.~O.} \emph{et~al.}
\newblock \bibinfo{title}{The earth surface mineral dust source investigation: An earth science imaging spectroscopy mission}.
\newblock \emph{\bibinfo{journal}{2020 IEEE Aerospace Conference}} \bibinfo{pages}{1--15} (\bibinfo{year}{2020}).

\bibitem{cawsenicholson2021nasa}
\bibinfo{author}{Cawse-Nicholson, K.} \emph{et~al.}
\newblock \bibinfo{title}{Nasa's surface biology and geology designated observable: A perspective on surface imaging algorithms}.
\newblock \emph{\bibinfo{journal}{Remote Sensing of Environment}} \textbf{\bibinfo{volume}{257}}, \bibinfo{pages}{112349} (\bibinfo{year}{2021}).
\newblock \urlprefix\url{https://www.sciencedirect.com/science/article/pii/S0034425721000675}.

\bibitem{nieke2023chime}
\bibinfo{author}{Nieke, J.} \emph{et~al.}
\newblock \bibinfo{title}{{The copernicus hyperspectral imaging mission for the environment (CHIME): an overview of its mission, system and planning status}}.
\newblock \emph{\bibinfo{journal}{Sensors, Systems, and Next-Generation Satellites XXVII}} \textbf{\bibinfo{volume}{12729}}, \bibinfo{pages}{1272909} (\bibinfo{year}{2023}).
\newblock \urlprefix\url{https://doi.org/10.1117/12.2679977}.

\bibitem{zhu2012object}
\bibinfo{author}{Zhu, Z.} \& \bibinfo{author}{Woodcock, C.~E.}
\newblock \bibinfo{title}{Object-based cloud and cloud shadow detection in landsat imagery}.
\newblock \emph{\bibinfo{journal}{Remote Sensing of Environment}} \textbf{\bibinfo{volume}{118}}, \bibinfo{pages}{83--94} (\bibinfo{year}{2012}).
\newblock \urlprefix\url{https://www.sciencedirect.com/science/article/pii/S0034425711003853}.

\bibitem{qiu2019fmask4}
\bibinfo{author}{Qiu, S.}, \bibinfo{author}{Zhu, Z.} \& \bibinfo{author}{He, B.}
\newblock \bibinfo{title}{Fmask 4.0: Improved cloud and cloud shadow detection in landsats 4–8 and sentinel-2 imagery}.
\newblock \emph{\bibinfo{journal}{Remote Sensing of Environment}} \textbf{\bibinfo{volume}{231}}, \bibinfo{pages}{111205} (\bibinfo{year}{2019}).
\newblock \urlprefix\url{https://www.sciencedirect.com/science/article/pii/S0034425719302172}.

\bibitem{yang2019feature}
\bibinfo{author}{Yang, J.} \emph{et~al.}
\newblock \bibinfo{title}{Cdnet: Cnn-based cloud detection for remote sensing imagery}.
\newblock \emph{\bibinfo{journal}{IEEE Transactions on Geoscience and Remote Sensing}} \textbf{\bibinfo{volume}{57}}, \bibinfo{pages}{6195--6211} (\bibinfo{year}{2019}).

\bibitem{hagolle2010multi}
\bibinfo{author}{Hagolle, O.}, \bibinfo{author}{Huc, M.}, \bibinfo{author}{Pascual, D.~V.} \& \bibinfo{author}{Dedieu, G.}
\newblock \bibinfo{title}{A multi-temporal method for cloud detection, applied to formosat-2, ven$\mu$s, landsat and sentinel-2 images}.
\newblock \emph{\bibinfo{journal}{Remote Sensing of Environment}} \textbf{\bibinfo{volume}{114}}, \bibinfo{pages}{1747--1755} (\bibinfo{year}{2010}).

\bibitem{zhu2014automated}
\bibinfo{author}{Zhu, Z.} \& \bibinfo{author}{Woodcock, C.~E.}
\newblock \bibinfo{title}{Automated cloud, cloud shadow, and snow detection in multitemporal landsat data: An algorithm designed specifically for monitoring land cover change}.
\newblock \emph{\bibinfo{journal}{Remote Sensing of Environment}} \textbf{\bibinfo{volume}{152}}, \bibinfo{pages}{217--234} (\bibinfo{year}{2014}).
\newblock \urlprefix\url{https://www.sciencedirect.com/science/article/pii/S0034425714002259}.

\bibitem{zhu2018automatic}
\bibinfo{author}{Zhu, X.} \& \bibinfo{author}{Helmer, E.~H.}
\newblock \bibinfo{title}{An automatic method for screening clouds and cloud shadows in optical satellite image time series in cloudy regions}.
\newblock \emph{\bibinfo{journal}{Remote Sensing of Environment}} \textbf{\bibinfo{volume}{214}}, \bibinfo{pages}{135--153} (\bibinfo{year}{2018}).
\newblock \urlprefix\url{https://www.sciencedirect.com/science/article/pii/S0034425718302530}.

\bibitem{green1998imaging}
\bibinfo{author}{Green, R.~O.} \emph{et~al.}
\newblock \bibinfo{title}{Imaging spectroscopy and the airborne visible/infrared imaging spectrometer (aviris)}.
\newblock \emph{\bibinfo{journal}{Remote Sensing of Environment}} \textbf{\bibinfo{volume}{65}}, \bibinfo{pages}{227--248} (\bibinfo{year}{1998}).
\newblock \urlprefix\url{https://www.sciencedirect.com/science/article/pii/S0034425798000649}.

\bibitem{gao1991cloud}
\bibinfo{author}{Gao, B.-C.} \& \bibinfo{author}{Goetz, A. F.~H.}
\newblock \bibinfo{title}{Cloud area determination from aviris data using water vapor channels near 1 \textmu m}.
\newblock \emph{\bibinfo{journal}{Journal of Geophysical Research: Atmospheres}} \textbf{\bibinfo{volume}{96}}, \bibinfo{pages}{2857--2864} (\bibinfo{year}{1991}).
\newblock \urlprefix\url{https://agupubs.onlinelibrary.wiley.com/doi/abs/10.1029/90JD02394}.

\bibitem{sun2020satellite}
\bibinfo{author}{Sun, L.} \emph{et~al.}
\newblock \bibinfo{title}{Satellite data cloud detection using deep learning supported by hyperspectral data}.
\newblock \emph{\bibinfo{journal}{International Journal of Remote Sensing}} \textbf{\bibinfo{volume}{41}}, \bibinfo{pages}{1349--1371} (\bibinfo{year}{2020}).

\bibitem{giuffrida2020cloudscout}
\bibinfo{author}{Giuffrida, G.} \emph{et~al.}
\newblock \bibinfo{title}{Cloudscout: A deep neural network for on-board cloud detection on hyperspectral images}.
\newblock \emph{\bibinfo{journal}{Remote Sensing}} \textbf{\bibinfo{volume}{12}} (\bibinfo{year}{2020}).
\newblock \urlprefix\url{https://www.mdpi.com/2072-4292/12/14/2205}.

\bibitem{zhai2018cloud}
\bibinfo{author}{Zhai, H.}, \bibinfo{author}{Zhang, H.}, \bibinfo{author}{Zhang, L.} \& \bibinfo{author}{Li, P.}
\newblock \bibinfo{title}{Cloud/shadow detection based on spectral indices for multi/hyperspectral optical remote sensing imagery}.
\newblock \emph{\bibinfo{journal}{ISPRS Journal of Photogrammetry and Remote Sensing}} \textbf{\bibinfo{volume}{144}}, \bibinfo{pages}{235--253} (\bibinfo{year}{2018}).
\newblock \urlprefix\url{https://www.sciencedirect.com/science/article/pii/S0924271618301989}.

\bibitem{thompson2014rapid}
\bibinfo{author}{Thompson, D.~R.} \emph{et~al.}
\newblock \bibinfo{title}{Rapid spectral cloud screening onboard aircraft and spacecraft}.
\newblock \emph{\bibinfo{journal}{IEEE Transactions on Geoscience and Remote Sensing}} \textbf{\bibinfo{volume}{52}}, \bibinfo{pages}{6779--6792} (\bibinfo{year}{2014}).

\bibitem{sandford2020global}
\bibinfo{author}{Sandford, M.~W.} \emph{et~al.}
\newblock \bibinfo{title}{Global cloud property models for real-time triage on board visible--shortwave infrared spectrometers}.
\newblock \emph{\bibinfo{journal}{Atmospheric Measurement Techniques}} \textbf{\bibinfo{volume}{13}}, \bibinfo{pages}{7047--7057} (\bibinfo{year}{2020}).
\newblock \urlprefix\url{https://amt.copernicus.org/articles/13/7047/2020/}.

\bibitem{thompson2024atbd}
\bibinfo{author}{Thompson, D.~R.} \emph{et~al.}
\newblock \bibinfo{title}{Emit l2a algorithm: Surface reflectance and scence content masks, theoretical basis}.
\newblock \emph{\bibinfo{journal}{Jet Propulsion Laboratory}}  (\bibinfo{year}{2024}).

\bibitem{gao2002algorithm}
\bibinfo{author}{Gao, B.-C.}, \bibinfo{author}{Yang, P.}, \bibinfo{author}{Han, W.}, \bibinfo{author}{Li, R.-R.} \& \bibinfo{author}{Wiscombe, W.~J.}
\newblock \bibinfo{title}{An algorithm using visible and 1.38-/spl mu/m channels to retrieve cirrus cloud reflectances from aircraft and satellite data}.
\newblock \emph{\bibinfo{journal}{IEEE Transactions on Geoscience and Remote Sensing}} \textbf{\bibinfo{volume}{40}}, \bibinfo{pages}{1659--1668} (\bibinfo{year}{2002}).

\bibitem{lundberg2017shap}
\bibinfo{author}{Lundberg, S.~M.} \& \bibinfo{author}{Lee, S.-I.}
\newblock \bibinfo{editor}{Guyon, I.} \emph{et~al.} (eds) \emph{\bibinfo{title}{A unified approach to interpreting model predictions}}.
\newblock (eds \bibinfo{editor}{Guyon, I.} \emph{et~al.}) \emph{\bibinfo{booktitle}{Advances in Neural Information Processing Systems}}, Vol.~\bibinfo{volume}{30} (\bibinfo{publisher}{Curran Associates, Inc.}, \bibinfo{year}{2017}).

\bibitem{qi2019ig}
\bibinfo{author}{Qi, Z.}, \bibinfo{author}{Khorram, S.} \& \bibinfo{author}{Li, F.}
\newblock \bibinfo{title}{Visualizing deep networks by optimizing with integrated gradients.}
\newblock \emph{\bibinfo{journal}{CVPR workshops}} \textbf{\bibinfo{volume}{2}}, \bibinfo{pages}{1--4} (\bibinfo{year}{2019}).

\bibitem{gao1993cirrus}
\bibinfo{author}{Gao, B.-C.}, \bibinfo{author}{Goetz, A.~F.} \& \bibinfo{author}{Wiscombe, W.~J.}
\newblock \bibinfo{title}{Cirrus cloud detection from airborne imaging spectrometer data using the 1.38 $\mu$m water vapor band}.
\newblock \emph{\bibinfo{journal}{Geophysical Research Letters}} \textbf{\bibinfo{volume}{20}}, \bibinfo{pages}{301--304} (\bibinfo{year}{1993}).

\bibitem{taylor2011comparison}
\bibinfo{author}{Taylor, T.~E.} \emph{et~al.}
\newblock \bibinfo{title}{Comparison of cloud-screening methods applied to gosat near-infrared spectra}.
\newblock \emph{\bibinfo{journal}{IEEE Transactions on Geoscience and Remote Sensing}} \textbf{\bibinfo{volume}{50}}, \bibinfo{pages}{295--309} (\bibinfo{year}{2011}).

\bibitem{pasquarella2023csplus}
\bibinfo{author}{Pasquarella, V.~J.}, \bibinfo{author}{Brown, C.~F.}, \bibinfo{author}{Czerwinski, W.} \& \bibinfo{author}{Rucklidge, W.~J.}
\newblock \bibinfo{title}{Comprehensive quality assessment of optical satellite imagery using weakly supervised video learning}.
\newblock \emph{\bibinfo{journal}{Proceedings of the IEEE/CVF Conference on Computer Vision and Pattern Recognition}} \bibinfo{pages}{2125--2135} (\bibinfo{year}{2023}).

\bibitem{dronner2018fast}
\bibinfo{author}{Dr{\"o}nner, J.} \emph{et~al.}
\newblock \bibinfo{title}{Fast cloud segmentation using convolutional neural networks}.
\newblock \emph{\bibinfo{journal}{Remote Sensing}} \textbf{\bibinfo{volume}{10}}, \bibinfo{pages}{1782} (\bibinfo{year}{2018}).

\bibitem{zhang2022cloudvit}
\bibinfo{author}{Zhang, B.}, \bibinfo{author}{Zhang, Y.}, \bibinfo{author}{Li, Y.}, \bibinfo{author}{Wan, Y.} \& \bibinfo{author}{Yao, Y.}
\newblock \bibinfo{title}{Cloudvit: A lightweight vision transformer network for remote sensing cloud detection}.
\newblock \emph{\bibinfo{journal}{IEEE Geoscience and Remote Sensing Letters}} \textbf{\bibinfo{volume}{20}}, \bibinfo{pages}{1--5} (\bibinfo{year}{2022}).

\bibitem{hong2021spectralformer}
\bibinfo{author}{Hong, D.} \emph{et~al.}
\newblock \bibinfo{title}{Spectralformer: Rethinking hyperspectral image classification with transformers}.
\newblock \emph{\bibinfo{journal}{IEEE Transactions on Geoscience and Remote Sensing}} \textbf{\bibinfo{volume}{60}}, \bibinfo{pages}{1--15} (\bibinfo{year}{2021}).

\bibitem{he2021spatial}
\bibinfo{author}{He, X.}, \bibinfo{author}{Chen, Y.} \& \bibinfo{author}{Lin, Z.}
\newblock \bibinfo{title}{Spatial-spectral transformer for hyperspectral image classification}.
\newblock \emph{\bibinfo{journal}{Remote Sensing}} \textbf{\bibinfo{volume}{13}}, \bibinfo{pages}{498} (\bibinfo{year}{2021}).

\bibitem{calef2024mmgis}
\bibinfo{author}{Calef, F.~J.} \& \bibinfo{author}{Soliman, T.~K.}
\newblock \bibinfo{title}{Nasa-ammos mmgis open source software}.
\newblock \emph{\bibinfo{journal}{GitHub}}  (\bibinfo{year}{2024}).
\newblock \urlprefix\url{https://github.com/NASA-AMMOS/MMGIS}.

\bibitem{chen2016xgboost}
\bibinfo{author}{Chen, T.} \& \bibinfo{author}{Guestrin, C.}
\newblock \bibinfo{title}{Xgboost: A scalable tree boosting system}.
\newblock \emph{\bibinfo{journal}{Proceedings of the 22nd acm sigkdd international conference on knowledge discovery and data mining}} \bibinfo{pages}{785--794} (\bibinfo{year}{2016}).

\bibitem{paszke2019pytorch}
\bibinfo{author}{Paszke, A.} \emph{et~al.}
\newblock \bibinfo{title}{Pytorch: An imperative style, high-performance deep learning library}.
\newblock \emph{\bibinfo{journal}{Advances in neural information processing systems}} \textbf{\bibinfo{volume}{32}} (\bibinfo{year}{2019}).

\bibitem{he2016deep}
\bibinfo{author}{He, K.}, \bibinfo{author}{Zhang, X.}, \bibinfo{author}{Ren, S.} \& \bibinfo{author}{Sun, J.}
\newblock \bibinfo{title}{Deep residual learning for image recognition}.
\newblock \emph{\bibinfo{journal}{Proceedings of the IEEE conference on computer vision and pattern recognition}} \bibinfo{pages}{770--778} (\bibinfo{year}{2016}).

\bibitem{defazio2024road}
\bibinfo{author}{Defazio, A.} \emph{et~al.}
\newblock \bibinfo{title}{The road less scheduled} (\bibinfo{year}{2024}).
\newblock \eprint{2405.15682}.

\bibitem{vaswani2017transformer}
\bibinfo{author}{Vaswani, A.} \emph{et~al.}
\newblock \bibinfo{editor}{Guyon, I.} \emph{et~al.} (eds) \emph{\bibinfo{title}{Attention is all you need}}.
\newblock (eds \bibinfo{editor}{Guyon, I.} \emph{et~al.}) \emph{\bibinfo{booktitle}{Advances in Neural Information Processing Systems}}, Vol.~\bibinfo{volume}{30} (\bibinfo{publisher}{Curran Associates, Inc.}, \bibinfo{year}{2017}).
\newblock \urlprefix\url{https://proceedings.neurips.cc/paper_files/paper/2017/file/3f5ee243547dee91fbd053c1c4a845aa-Paper.pdf}.

\bibitem{dosovitskiy2021vit}
\bibinfo{author}{Dosovitskiy, A.} \emph{et~al.}
\newblock \bibinfo{title}{An image is worth 16x16 words: Transformers for image recognition at scale}.
\newblock \emph{\bibinfo{journal}{International Conference on Learning Representations}}  (\bibinfo{year}{2021}).

\bibitem{kenton2019bert}
\bibinfo{author}{Kenton, J. D. M.-W.~C.} \& \bibinfo{author}{Toutanova, L.~K.}
\newblock \bibinfo{title}{Bert: Pre-training of deep bidirectional transformers for language understanding}.
\newblock \emph{\bibinfo{journal}{Proceedings of naacL-HLT}} \textbf{\bibinfo{volume}{1}}, \bibinfo{pages}{2} (\bibinfo{year}{2019}).

\bibitem{liu2021exploring}
\bibinfo{author}{Liu, S.}, \bibinfo{author}{Le, F.}, \bibinfo{author}{Chakraborty, S.} \& \bibinfo{author}{Abdelzaher, T.}
\newblock \bibinfo{title}{On exploring attention-based explanation for transformer models in text classification}.
\newblock \emph{\bibinfo{journal}{2021 IEEE International Conference on Big Data (Big Data)}} \bibinfo{pages}{1193--1203} (\bibinfo{year}{2021}).

\end{thebibliography}

\end{document}